\documentclass[letterpaper, 10 pt, conference]{ieeeconf}  

\IEEEoverridecommandlockouts                              

\usepackage{graphicx}
\usepackage{url}  
\usepackage{amsmath}
\usepackage{amssymb}
\usepackage{multirow}
\usepackage[table,xcdraw]{xcolor}

\title{\LARGE \bf
LSTM-based Deep Neural Network With A Focus on Sentence Representation for Sequential Sentence Classification \\ in Medical Scientific Abstracts  
}

\author{Phat~Lam$^{1*}$, 
        Lam~Pham$^{2*}$,
        Tin~Nguyen$^{3}$, 
        Hieu~Tang$^{4}$,
        Michael~Seidl$^{5}$,
        Medina~Andresel$^{6}$,
        Alexander~Schindler$^{7}$  
\thanks{L. Pham, M. Seidl, M. Andresel and A. Schindler are with Austrian Institute of Technology, Austria.}%
\thanks{H. Tang is with FPT University, Vietnam.}%
\thanks{P. Lam and T. Nguyen are with HCM University of Technology, Vietnam.}%
\thanks{(*) Main and equal contribution into the paper.}
}

\begin{document}

\maketitle
\thispagestyle{empty}
\pagestyle{empty}

\begin{abstract}
The Sequential Sentence Classification task within
the domain of medical abstracts, termed as SSC, involves the
categorization of sentences into pre-defined headings based on
their roles in conveying critical information in the abstract. In
the SSC task, sentences are sequentially related to each other.
For this reason, the role of sentence embeddings is crucial for
capturing both the semantic information between words in the
sentence and the contextual relationship of sentences within the
abstract, which then enhances the SSC system performance. In this paper, we propose a LSTM-based deep learning network with a focus on creating comprehensive sentence representation at the sentence level. To
demonstrate the efficacy of the created sentence representation,  
a system utilizing these sentence embeddings is also developed, which consists of a Convolutional-Recurrent neural network (C-RNN) at the abstract level and a multi-layer perception network (MLP) at the segment level. Our proposed system yields highly competitive results compared to state-of-the-art systems and further enhances the F1 scores of the baseline by 1.0\%, 2.8\%, and 2.6\% on the benchmark datasets PudMed 200K RCT, PudMed 20K RCT and NICTA-PIBOSO, respectively. This indicates the significant impact of improving sentence representation on boosting model performance.

\end{abstract}
\section{INTRODUCTION}
\label{intro}
When doing research on a large-scale source of scientific papers, it is necessary to skim through abstracts to identify whether papers align with the research interest. 
This process becomes more straightforward when abstracts are organized with semantic headings such as "background", "objective", "methods", "results", and "conclusion". 
Therefore, automatically categorizing each sentence in a scientific abstract into a relevant heading, known as the task of Sequential Sentence Classification (SSC), significantly facilitates the information retrieval process within large-scale data. 
In medical domain, research abstracts present a large volume and have grown exponentially. 
Manually sorting through these documents to find relevant insights presents a time-consuming and labor-intensive task.
Therefore, the result of the SSC tasks enables researchers and learners to catch up and categorize research abstracts effectively.
In other words, the SSC task significantly facilitates learners and researchers by accelerating their educational processes of literature review, information extraction, evidence-based decision-making, etc.  
Recently, the SSC task in medical scientific abstracts has drawn attention from NLP research community.
Indeed, some large and benchmark datasets such as PubMed RCT~\cite{pubmed_ds} and NICTA-PIBOSO~\cite{nicta} were published.
Additionally, a wide range of machine learning based and deep learning based models have been proposed for this task.
Traditional machine learning methods utilized hand-crafted feature extraction for individual sentences. These extracted features are related to lexical, semantic, and structural information of an individual sentence such as synonyms, bag-of-words, part of speech, etc.
Then, sentences are classified by Hidden Markov Model (HMM)~\cite{hhm2006}, Naive Bayes~\cite{NaiveBY-2007} or CRF~\cite{crf-kim}. 
While the traditional machine learning based models present a limitation of exploring the relation among the sentences as using the hand-crafted features, leveraging deep neural networks in deep learning based models allows to capture the patterns of contextual relationship among sentences in the same abstract that leads to a breakthrough on model performance. 
For example, Dernoncourt et al.~\cite{bi-ann} introduced an deep learning model which uses a CRF layer to optimize the predicted label sequence, where adjacent sentences have an impact on the prediction of each other. 
Jin and Szolovits~\cite{HSLN} proposed a hierarchical sequential labeling network to further improve the semantic information within surrounding sentences for classification. 
Recently, Yamada et al.~\cite{yamada-93.1} and Shang et al.~\cite{shang-92.8} introduced some methodologies to assign labels to span sequences at the span level, which achieved state-of-the-art results. However, these two systems consider all possible span sequences of various lengths, which is very computationally expensive on large datasets. Importantly, these systems attempted to analyze the sequence of sentences at the span level without initially considering the improvement of the sentence representation, which is the fundamental components of this specific SSC task.  
At the sentence level, these systems leveraged pre-trained m BERT model~\cite{blue-bert} for the medical domain, which is trained on various NLP tasks, to extract the sentence embeddings. Commonly, BERT model primarily focuses on capturing syntactic meaning and contextual dependencies of words within individual sentences or pairs of sentences \cite{org_bert}. To some extent, the extracted embeddings may lack of the ability to grasp dependencies between sentences in a wider context (e.g. abstracts, documents), which is one of the important task-specific properties of the SSC task. 

%

In this paper, we aim to improve the sentence representation and explore its impact on the performance of SSC task in medical scientific abstracts. We propose a deep neural network with a focus on extracting well-presented sentence embeddings.
In particular, we explore the independent features of sentence, word sequence, character sequence, and statistic information of sentences in one abstract. Then, we develop a LSTM-based deep neural network with multiple-feature branches for classifying individual sentences.
The network is then used to extract the comprehensive sentence embeddings.
Given these sentence embeddings, a Convolutional-RNN based network (C-RNN) at the abstract level and a Multi-layer Perception network (MLP) at the segment level are introduced to learn the contextual patterns of sentences in the same abstract. 
Finally, the results of C-RNN and MLP models are fused to achieve the final predicted sentences in an abstract.
We evaluate our proposed models on two benchmark datasets, PubMed RCT~\cite{pubmed_ds} and NICTA-PIBOSO~\cite{nicta}.
The experimental results indicate that exploiting multiple features extracted from sentences such as word sequence, character sequence, and statistic information of sentences in the abstract helps to generate well-presented sentence embeddings at the sentence level.
Both C-RNN network at the abstract level and MLP network at the segment level respectively further improve the performance when leveraging these well-presented sentence embeddings. 

\section{The Overall Proposed System}
\label{our_system}
\begin{figure}[t]
    	\vspace{-0.2cm}
    \centering
    \includegraphics[width =1\linewidth]{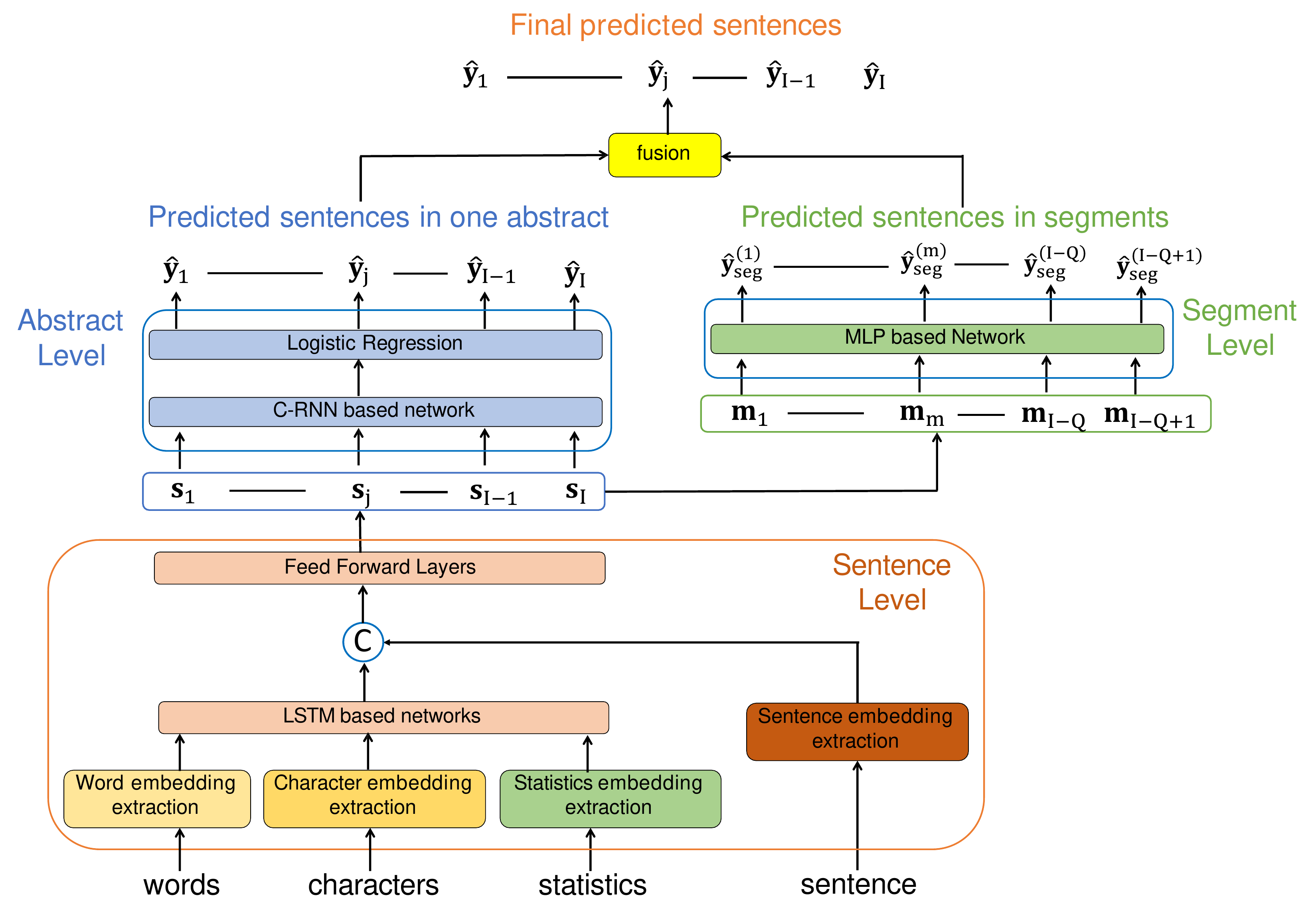}
	\caption{The overall architecture of the proposed system}
   	\vspace{-0.4cm}
    \label{fig:f1}
\end{figure}

The proposed system in this paper for the task of sequential sentence classification in medical scientific abstracts is generally presented in Fig.~\ref{fig:f1}.
As Fig.~\ref{fig:f1} shows, the proposed network comprises of three main sub-networks, referred to as the classification model (Sen-Model) at the sentence level, the regression model at the abstract level (Abs-Model) and the classification model at the segment level (Seg-Model). 
At the sentence level, we establish the task sentence classification for individual sentences. 
The proposed LSTM-based classification model at the sentence level (Sen-Model) presents 4 branches, each of which explores the distinct feature from the full sentence, the words in the sentence, the character in the sentence, and the statistical information of the sentence in one abstract, aiming to achieve the comprehensive and specific sentence representation.
Given the classification model at the sentence level, we extract the sentence embeddings $\mathbf{S} = [\mathbf{s}_1, \mathbf{s}_2, ..., \mathbf{s}_S]$, where $\mathbf{s}_s$ represents an individual sentence.
The sentence embeddings are then utilized in the regression model at the abstract level (Abs-Model) and the classification model at the segment level (Seg-Model) to further improve the tasks of sentence classification by exploiting the properties of the well-presented sentence representation at higher 
contextual levels.
Both the classification model at the sentence level and the regression model at the abstract level leverage the RNN-based architecture, the attention mechanism, and the multilayer perception (MLP) architecture which are comprehensively presented in next sections.
\subsection{The classification model at the sentence level (Sen-Model)}

\label{sentence}
\begin{figure}[t]
    	\vspace{-0.2cm}
    \centering
    \includegraphics[width =0.95\linewidth]{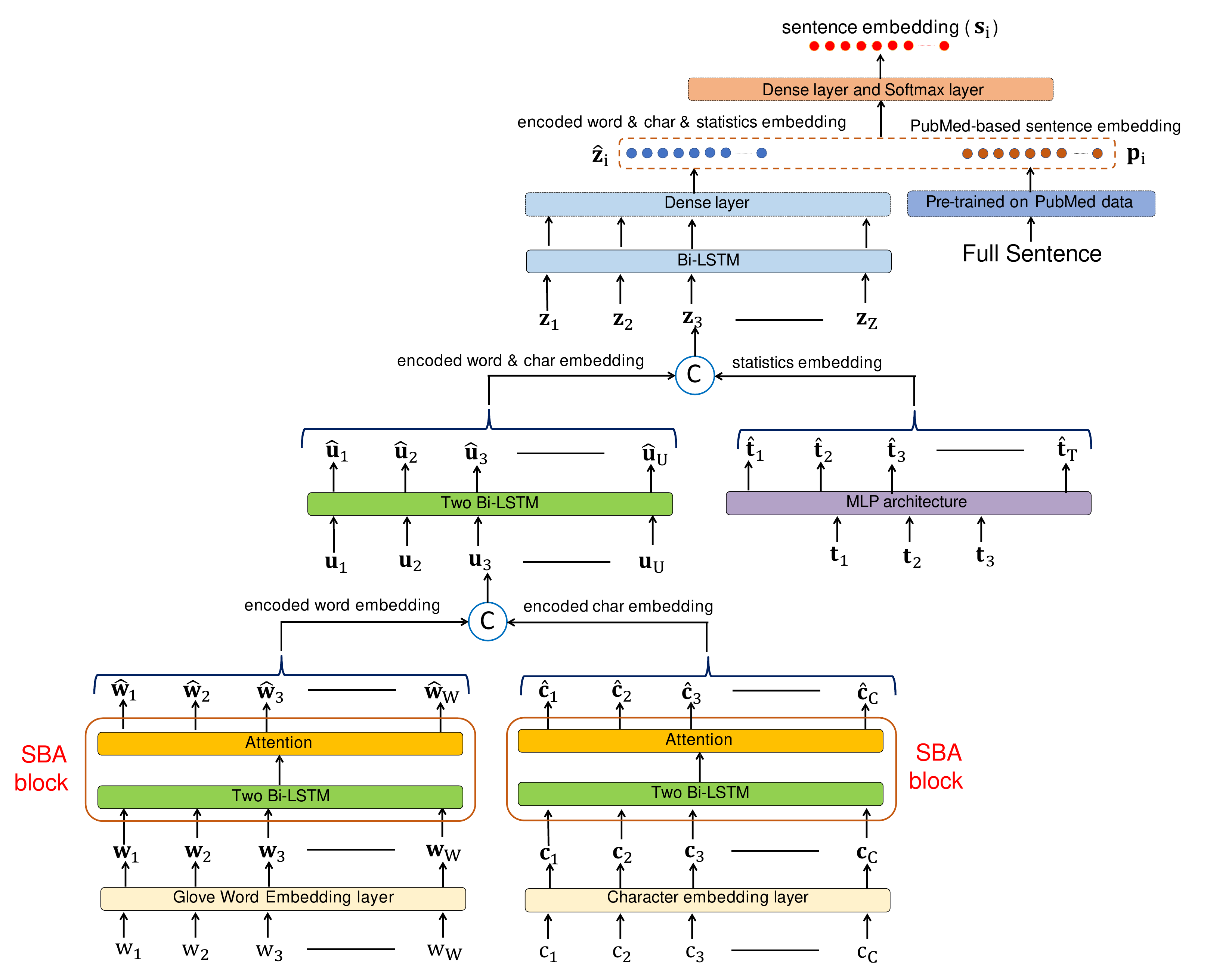}
       	\vspace{-0.2cm}
	\caption{The Sen-Model architecture for classification at the sentence level}
   	\vspace{-0.4cm}
    \label{fig:f2}
\end{figure}

The proposed LSTM-based network focus on improving sentence representation at the sentence level (Sen-Model) is comprehensively presented in Fig.~\ref{fig:f2}.  
Given a sentence including $W$ words $[w_1, w_2, ... w_{W}]$ and $C$ characters $[c_1, c_2, ... c_{C}]$.
We make use of the Glove~\cite{glove} model to extract a sequence of word embeddings $\mathbf{W} = [\mathbf{w}_1, \mathbf{w}_2, ... \mathbf{w}_{W}]$, where $\mathbf{w}_{w} \in \mathbb{R}^{d_{w}}$ presents a word embedding and $d_{w}$ is the dimension of a word embedding.
Regarding the sequence of characters in one sentence, the character embedding is randomly initialized in the uniform distribution to extract the character embeddings $\mathbf{C} = [\mathbf{c}_1, \mathbf{c}_2, ... \mathbf{c}_{C}]$, where $\mathbf{c}_{c} \in \mathbb{R}^{d_{c}}$ presents a character embedding and $d_{c}$ is the dimension of a character embedding.
%
%
The sequence of word embeddings $\mathbf{W}$ and the sequence of character embeddings $\mathbf{C}$ are fed into stacked Bi-LSTM-Attention encoder blocks, referred as SBA blocks, to generate the encoded word embeddings $\widehat{\mathbf{W}} = [\widehat{\mathbf{w}}_1, \widehat{\mathbf{w}}_2, ... \widehat{\mathbf{w}}_W]$ and the encoded character embeddings $\widehat{\mathbf{C}} = [\widehat{\mathbf{c}}_1, \widehat{\mathbf{c}}_2, ... \widehat{\mathbf{c}}_C]$, where $\widehat{\mathbf{w}}_w,\, \widehat{\mathbf{c}}_c \in \mathbb{R}^{d_h}$ with $h$ is the hidden state dimension. 
The SBA block includes a Bi-LSTM network which comprises of two stacked Bidirectional LSTM layers, followed by a Scaled Dot-Product Attention layer~\cite{attention}. 
Each Bidirectional LSTM layer takes the output sequence of the previous layer as input, which allows the capture of more complex lexical, syntactic, and semantic information between words and characters in an individual sentence. 
Given the sequential word representation and the sequential character representation extracted from the Bidirectional LSTM layers, we apply linear transform to create query, key and value matrix $\mathbf{Q}\in R^{N_q\times d_l}$, $\mathbf{K}\in R^{N_k\times d_l}$, $\mathbf{V}\in R^{N_v\times d_v}$, where $N_q$, $N_k$, $N_v$ are the number of queries, keys and values, $d_l$ and $d_v$ are the dimension of query and key, the dimension of value, respectively. The output matrix of the Scaled Dot-Product attention layer is computed as:
\begin{equation}
   Attention(\mathbf{Q, K, V})=\text{Softmax}\left(\frac{\mathbf{QK}^T}{\sqrt{d_l}}\right)\mathbf{V}
\end{equation}

Two encoded embeddings $\widehat{\mathbf{W}}$ and $\widehat{\mathbf{C}}$  extracted from SBA blocks of words and characters are concatenated to generated the word-char embedding $\mathbf{U} = [\mathbf{u}_1, \mathbf{u}_2, ... \mathbf{u}_U]$ where $U = W + C$ and $ \mathbf{u}_u \in \mathbb{R}^{d_h}$. The word-char embedding $\mathbf{U}$ is then fed into the word-char encoder block to generate the encoded word-char embeddings $\widehat{\mathbf{U}}= [\widehat{\mathbf{u}}_1, \widehat{\mathbf{u}}_2, ... \widehat{\mathbf{u}}_U]$. The word-char encoder block reuses the two stacked Bidirectional LSTM layers from the SBA block without using the attention layer.

Besides the lexical, syntactic and semantic information for each sentence extracted from the word and character branches, we consider the statistical information of individual sentence: the number of sentences in the same abstract, the index of sentence in the abstract, and the number of words in the sentence, which are represented by one-hot vectors $\mathbf{t_1, t_2, t_3}$. The statistical information equips each sentence with the ability to capture sequential and contextual properties related to other sentences within abstract. The statistical vectors are fed into a Multi-layer perception (MLP) 
to generate encoded statistic embeddings $\widehat{\mathbf{T}} = [\widehat{\mathbf{t}}_1, \widehat{\mathbf{t}}_2, ... \widehat{\mathbf{t}}_T]$.
%
%
%
%
The encoded statistic embeddings $\widehat{\mathbf{T}}$ are then concatenated with the encoded word-char embedding $\widehat{\mathbf{U}}$ to generate the word-char-stat embedding $\mathbf{Z} = [\mathbf{z}_1, \mathbf{z}_2, ... \mathbf{z}_Z]$ where $Z = U + T$ and $\mathbf{z}_z \in \mathbb{R}^{d_h}$.
Again, one Bi-LSTM layer and one Dense layer are used to learn the sequence of word-char-stat embeddings, which combines both semantic, syntactic information of word-char encoded embedding and statistical information of statistic embedding, to generate the encoded word-char-stat embeddings $\widehat{\mathbf{Z}}$.

To further enhance the representation of sentences in term of language comprehension specifically in biomedical domains, we utilize BiomedBERT~\cite{BiomedBERT}, which was pretrained on the PubMed corpus. 
The PubMed-based sentence embedding $\mathbf{P} = [\mathbf{P}_1, \mathbf{P}_2, ... \mathbf{P}_P]$ is concatenated with the encoded word-char-stat embedding $\widehat{\mathbf{Z}}$ before feeding into a Dense layer followed by a Softmax layer for classification.
After training the Sen-Model at the sentence level, for each sentence, we extract the output of the Dense layer before the final Softmax layer and consider it as the final sentence-level embedding $\mathbf{s}_i \in R^{d_{L}}$, where $d_{L}$ is the dimension of a sentence-level embedding (i.e. the Softmax layer presents $L$ outputs which match $L$ labels of sentences in an abstract). The final sentence representation of the entire dataset is $\mathbf{S} = [\mathbf{s}_1, \mathbf{s}_2, ... \mathbf{s}_S]$, where $S$ is the number of sentences in the dataset and $\mathbf{s}_i \in R^{d_{L}}$. The Sen-Model is optimized using Categorical Cross-Entropy:
\begin{equation}
    L_{\text{Sen}} = -\frac{1}{N}\sum^{N}_{i=1}\sum^{L}_{j=1}y_{ij}\log{\widehat{y}_{ij}}
\end{equation}
where $\mathbf{y}_i$, $\widehat{\mathbf{y}}_i$ and $N$ are the true label, the predicted probability vector of sentence $\mathbf{s}_i$ and the batch number, respectively. To examine the efficacy of the extracted sentence embeddings, we constructed two networks aimed at enhancing performance at higher levels by leveraging these embeddings, which are presented in the next subsections.
\begin{table}[t]
 \caption{MLP based network for classification at the segment level} 
        	\vspace{-0.2cm}
    \centering
    \scalebox{0.83}{
\begin{tabular}{clc}
\hline
\textbf{Blocks} & \multicolumn{1}{c}{\textbf{Layers}} & \textbf{Output Shape} \\ \hline
F1                      & Dense (512) - Elu - BN -Dr(0.5)      & 512                   \\ \hline
F2                      & Dense (256) - Elu - BN - Dr(0.5)      & 256                   \\ \hline
F3                      & Dense (128) - Elu - BN - Dr(0.5)      & 128                   \\ \hline
F4                      & Dense (64) - Elu - BN - Dr(0.5)       & 64                    \\ \hline
F5                      & Dense (L) - Softmax        & L                     \\ \hline
\end{tabular}
    }
  	\vspace{-0.4cm}
\label{table_segment}
\end{table}
\subsection{The regression model at the abstract level (Abs-Model)}
\label{abtract}
Given the original dataset comprising of $S$ sentences $[s_1, s_2, ... s_S]$, each sentence is now represented by a sentence-level embedding $\mathbf{s}_i,\,i=1, 2, ..., S$,  extracted from the Sen-Model at the sentence level. 
To explore the sequential and contextual properties of sentences in one abstract, we group sentence embeddings in the same abstract to create the abstract representation $\mathbf{A} = [\mathbf{s}_1, \mathbf{s}_2, ... \mathbf{s}_I]$ where $\mathbf{s}_i \in \mathbb{R}^{d_{L}}$ and $I$ is the number of sentences in one abstract. 
The abstract representation $\mathbf{A}$ is a sequence of sentence-level embeddings which is fed into the regression model at the abstract level (Abs-Model).
The regression model at the abstract level (Abs-Model) is comprehensively presented at the left corner of the upper part of Fig.~\ref{fig:f1}. The network includes two parts:  Convolutional-Recurrent Neural Network (C-RNN) and Logistic Regression classifier.

%
%
Each abstract representation $\mathbf{A}$ is now considered as a two-dimensional tensor which is fed into the convolution layers to extract essential features represented for neighbour sentences in one abstract. 
The two 2D-convolution layers in the C-RNN present the same kernel size $(8, 3)$ and the same padding. 
The number of filters in each layer are set to $16$.
Next, the Bi-RNN decoder is used for learning sequential relationship feature maps extracted from the convolutional layers. 
Finally, the Logistic Regression classifier receives the feature maps from the Bi-RNN decoder as input and generate predicted values $\widehat{\mathbf{{Y}}}_{\text{abs}} = [\widehat{\mathbf{y}}_1,\widehat{\mathbf{y}}_2, ... \widehat{\mathbf{y}}_\text{I}]$ corresponding to the ground truth $\mathbf{{Y}}_{\text{abs}} = [\mathbf{y}_1, \mathbf{y}_2, ... {\mathbf{y}}_\text{I}]$ 
where $\mathbf{\widehat{y}}_i, \mathbf{\mathbf{y}}_i\in R^{d_{L}}$. Regarding the predicted value and the ground truth of one abstract, we form a predicted sequence 
$\widehat{\mathbf{y}}_{\text{abs}}\in \mathbb{R}^{d_L\times I}$ and  $\mathbf{y}_{\text{abs}}\in \mathbb{R}^{d_L\times I}$ by concatenating all the vectors of $\widehat{\mathbf{{Y}}}_{\text{abs}}$ and $\mathbf{{Y}}_{\text{abs}}$, respectively. Then, the Abs-Model is optimized using Binary Cross Entropy loss on these predicted sequences, which can be written as:
\begin{equation}
    L_{\text{Abs}}=-\frac{1}{N}\sum^{N}_{i=1}\sum^{d_L\times I}_{j=1}\left(y_{ij}\log(\widehat{y}_{ij})+ (1-y_{ij})\log(1-\widehat{y}_{ij})\right)
\end{equation}
where $N$ is the batch number and the innermost sum presents the BCE loss for one abstract.
\subsection{The classification model at the segment level (Seg-Model)}
Given the extracted sentence embeddings, instead of generating all the segments with various lengths, we create fixed-length segments with the size of $Q$ by grouping every $Q$ consecutive sentences in one abstract. Each abstract of $I$ sentences has $I-Q+1$ segments. The $i^{\text{th}}$ segment representation is described as $\mathbf{m}^{(i)}=[\mathbf{s}_{Qi}\, \mathbf{s}_{Qi+1}\, ...\, \mathbf{s}_{Qi+Q-1}]$, which is formed by concatenating $Q$ continuous sentence embeddings.
The corresponding label vector $\mathbf{y}_{\text{seq}}^{(i)}$ of the $i^{\text{th}}$  segment is defined as:
\begin{equation}
\mathbf{y}_{\text{seq}}^{(i)}=\frac{\displaystyle{\sum^{Qi+Q-1}_{q=Qi}\mathbf{y}_q}}{\displaystyle{\sum^{Qi+Q-1}_{q=Qi}\sum^{\text{L}}_{l=1} y_{ql}}}
\end{equation}                               



\noindent where $\displaystyle{\sum^{\text{L}}_{l=1} y_{ql}}$ is the sum of elements in the label vector $\mathbf{y}_q$ of the sentence $\mathbf{s}_{q}$. 
The fixed-length $Q$ is set to 3 based on empirical experiments.  
The Seg-model at the segment level uses the same labels as which of the sentence level, meaning that all the sentences in a segment receive the label of that segment. 

To classify segments, we use the MLP network which is shown in detail
at table \ref{table_segment}. The network consists of five fully-connected blocks. The first four blocks present the same layers which perform Dense layer, ELU activation, Batch Normalization and Dropout, respectively. The output of the last block is used for segment-based classification task. Since the labels of segment embeddings are no longer one-hot encoded, we use the Kullback-Leibler (KL) divergence loss for the segment-based classification task, which is defined as:
\begin{equation}
L_{\text{Seg}}(\theta) = \sum ^{N}_{n=1}\mathbf{y}_{\text{seq}}^{(n)}\log{\frac{\mathbf{y}_{\text{seq}}^{(n)}}{\widehat{\mathbf{y}}_{\text{seq}}^{(n)}}}+\frac{\lambda}{2}||\theta||^{2}_{2}    
\end{equation}

\noindent where $\theta$ is the trainable parameters of the network, $\lambda$ denotes the $l_2$ regularization coefficient experimentally set to 0.0001, $N$ is the batch number, $\mathbf{y}_{\text{seq}}^{(n)}$ and $\widehat{\mathbf{y}}_{\text{seq}}^{(n)}$ denote the ground-truth
and the network output in a batch, respectively.

\subsection{Inference with the entire system}
Given the predicted labels of Abs-Model at the abstract level and Seg-Model at the segment level, referred to as $\hat{\mathbf{Y}}_{\text{abs}}$ and $\hat{\mathbf{Y}}_{\text{seg}}$, the final predicted labels of our proposed system is defined as:
\begin{equation}
\hat{\mathbf{Y}}= \lambda_{\text{abs}}\hat{\mathbf{Y}}_{\text{abs}} +\lambda_{\text{seg}}\hat{\mathbf{Y}}_{\text{seg}}
\end{equation}
where $\lambda_{\text{abs}}$ and $\lambda_{\text{seg}}$ are the hyperparameters to control the predicted labels at the abstract level and the segment level.

\section{Experiment And Results}

\subsection{Datasets}
In this paper, we evaluate our proposed deep neural networks on two benchmark datasets: PubMed RCT~\cite{pubmed_ds} and NICTA-PIBOSO~\cite{nicta}. 

\textbf{PudMed RCT}: This dataset presents the largest and published dataset of text-based medical scientific abstracts. In particular, the PubMed dataset presents approximately 200,000 abstracts of randomized controlled trials.
Each sentence of each abstract is labeled with `BACKGROUND', `OBJECTIVE', `METHOD', `RESULT', or `CONCLUSION' which matches its role in the abstract.
The PubMed dataset proposed two sets of PubMed 20K and PubMed 200K, each of which presents three subsets of Training, Validation, and Test for training, validation and test processes, respectively.

\textbf{NICTA-PIBOSO}: This dataset is the official dataset of the
ALTA 2012 Shared Task.  The task was to build classifiers which  automatically divide sentences to a pre-defined set of categories in the domain of Evidence Based Medicine (EBM), which are 'BACKGROUND', 'INTERVENTION', 'OUTCOME', 'POPULATION', 'STUDY DESIGN', 'OTHER'.
\subsection{Evaluation metric}
\begin{table}[t]
    \caption{Compare our proposed systems with the baseline on the Test set (F1 score/Presision/Recall)} 
        	\vspace{-0.2cm}
    \centering
    \scalebox{0.83}{
    \begin{tabular}{l c c c} 
        \hline 
        \textbf{Systems}   &  \textbf{PubMed 20K}   & \textbf{NICTA-PIBOSO} \\       
        \hline
         bi-ANN~\cite{bi-ann} (baseline)      &90.0/-/- & 82.7/-/-\\
        \hline
         Sen-Model w/ word only    &84.0/84.2/83.9 &69.9/70.3/69.8\\
         Sen-Model w/ word \& char   &84.2/84.2/84.2  &70.0/70.3/69.8\\
         Sen-Model w/ word \& char \& stat     & 89.5/89.7/89.3  &77.9/77.9/77.9\\ 
         Sen-Model w/ pre-trained sentence only & 87.0/87.1/87.0  &78.5/78.8/78.5\\
         Sen-Model w/ sentence \& word \& char \& stat     &\textbf{91.1/91.9/90.9} &\textbf{81.8/81.8/81.8}\\ 
        \hline
         Abs-Model w/ word only     &90.6/91.2/90.4  &81.5/83.4/80.3\\ 
         Abs-Model w/ word \& char     &90.7/91.3/90.5  &81.4/83.1/80.5\\ 
         Abs-Model w/ word \& char \& stat     & 91.5/91.8/91.2 &81.2/82.6/80.1 \\ 
         Abs-Model w/ pre-trained sentence only & 91.9/92.1/91.7  &82.5/84.0/81.5\\
         Abs-Model w/ sentence \& word \& char \& stat     & \textbf{92.7/93.2/92.6} &\textbf{84.6/85.5/84.1}\\ 
       \hline 
    \end{tabular}
    }
    \label{table:res_01} 
\end{table}
\begin{table}[t]
     \caption{Compare Our Proposed Systems on Different Levels with the Baseline on The Test set of Two Benchmark Datasets(F1 score/Presision/Recall)} 
        	\vspace{-0.2cm}
    \centering
    \scalebox{0.9}{
    \begin{tabular}{l l l c} 
        \hline 
        \textbf{Systems}   &  \textbf{PubMed 20K}   & \textbf{NICTA-PIBOSO} \\       
        \hline
         bi-ANN~\cite{bi-ann} (baseline)      &90.0/-/- & 82.7/-/-\\
        \hline
        Sen-Model (Sentence) &91.1/91.9/90.9& 81.8/81.8/81.8\\ 
       Abs-Model  (Abstract)& 92.7/93.2/92.6
       &84.6/85.5/84.1\\ 
       Seg-Model  (Segment) & 91.0/92.5/89.6&79.5/80.7/78.5\\ 
       Combine-Model  & \textbf{92.8/93.4/92.7} &\textbf{85.3/86.5/84.5}\\ 
       \hline
    \end{tabular}}
    \label{res_01_01}
\end{table}
In this paper, we follow the original paper~\cite{pubmed_ds, nicta} which proposed the PubMed RCT and NICTA-PIBOSO datasets.
We then use Precision, Recall, and F1 scores as the evaluation metrics.
%
\subsection{Experimental settings}
We construct our proposed deep neural networks with the TensorFlow framework.
While the deep neural network used for Sen-Model is trained for 30 epochs, we train Abs-Model and Seg-Model with 60 epochs.
All deep neural networks in this paper are trained with the Titan RTX 24GB GPU. 
We use the Adam~\cite{Adam} method for the optimization. The learning rate for Sen-Model, Abs-Model and Seg-Model are 0.001, 0.003 and 0.001, respectively. 
A reduce learning rate scheme by a factor of 0.1 is set during training.
The Bi-RNN decoder at Abs-Model uses Bi-LSTM for PudMed dataset and Bi-GRU for NICTA-PIBOSO dataset, respectively. The hyperparameters $\lambda_{\text{abs}}$ and $\lambda_{\text{seg}}$ are empirically set to $1$ and $0.2$, respectively. 
TThe hidden states dimension $d_h$ of all LSTM layers in Sen-Model is set to 128, while the RNN layer at Abs-Model has 40 and 36 hidden states at two above datasets. 

\subsection{Experimental results}
\begin{table}[t]
    \caption{Compare our best model with the state-of-the-art systems On Test set of PudMed 20k dataset} 
    \centering
    \scalebox{0.85}{
\begin{tabular}{llc}
\hline
\multicolumn{1}{c}{\textbf{Authors}} & \multicolumn{1}{c}{\textbf{Systems}} & \multicolumn{1}{c}{\textbf{F1-score}} \\ \hline
Yamada et al.~\cite{yamada-93.1}     & Semi-Markov CRFs                    & \textbf{93.1}                                 \\ 
Athur Brack et al.~\cite{brack-93.0}    & Transfer /Multi-task learning       & 93.0                                  \\ 
Cohan et al.~\cite{cohan-92.9}        & Pretrained BERT                     & 92.9                                  \\ 
Xichen Shang et al.~\cite{shang-92.8}  & SDLA                                & 92.8                                  \\ 
Jin and Szolovits~\cite{HSLN}    & HSLN                                & 92.6                                  \\
Gaihong Yu et al.~\cite{Gaihong-91.15} & MSM & 91.2
    \\
Gonçalves et al.~\cite{goncaves-91.0}     & CNN-GRU                             & 91.0                                  \\
Dernoncourt et al.~\cite{bi-ann}   & bi-ANN                              & 90.0                                  \\ 
Agibetov et al.~\cite{agibetov-89.6}    & fastText                            & 89.6         \\                         
\hline
\textbf{Our proposed model}    & BiLSTM-CRNN-MLP                 & 92.8                                  \\ \hline
            
\end{tabular}
    }
    \label{table:res_02} 
\end{table}

\begin{table}[t]
    \caption{Compare our best model with the state-of-the-art systems on the Test set of NICTA-PIBOSO dataset} 
        	\vspace{-0.2cm}
    \centering
    \scalebox{0.85}{
\begin{tabular}{llc}
\hline
\multicolumn{1}{c}{\textbf{Authors}} & \multicolumn{1}{c}{\textbf{Systems}} & \multicolumn{1}{c}{\textbf{F1-score}} \\ \hline
Xichen Shang et al.~\cite{shang-92.8}  & SDLA                                & \textbf{86.8}                                  \\ 
Athur Brack et al.~\cite{brack-93.0}    & Transfer /Multi-task learning       & 86.0
\\
Yamada et al.~\cite{yamada-93.1}     & Semi-Markov CRFs                    & 84.4                                 \\  
Jin and Szolovits~\cite{HSLN}    & HSLN                                & 84.3 
\\  
Sarker et al.~\cite{sarker-nicta} & SVM &84.1
\\
Cohan et al.~\cite{cohan-92.9}        & Pretrained BERT                     & 83.0  \\
Dernoncourt et al.~\cite{bi-ann}   & bi-ANN                              & 82.7                                 \\ 
M Lui ~\cite{lui2012},~\cite{nicta2022overview}   & Feature stacking + Metalearner &   82.0   
    \\ \hline
\textbf{Our proposed model}    & BiLSTM-CRNN-MLP                  & 85.3                                 \\ \hline   
\end{tabular}
    }
    \vspace{-0.5cm}
    \label{table:res_03} 
\end{table}

We first evaluate our proposed models at the sentence level with different input features: using only word sequence (Sen-Model w/ word only); using both word and character sequences (Sen-Model w/ word \& char); using word, character, and statistics (Sen-Model w/ word \& char \& stat); using only sentence embeddings extracted from the pre-trained PudMed model (Sen-Model w/ sentence only); using all input features of word, character, statistics, sentence embeddings (Sen-Model w/ sentence \& word \& char \& stat).
The experimental results shown in the middle part of Table~\ref{table:res_01} prove that each input feature helps to further improve the performance at the sentence level.
The best performance at the sentence level is from the combination of all input features of word, character, statistics, sentence embeddings (Sen-Model w/ sentence \& word \& char \& stat), presenting the F1/Precision/Recall scores of 91.1/91.9/90.9 and 81.8/81.8/81.8 on PubMed 20K and NICTA-PIBOSO datasets, respectively. The model performance on the combination of multiple features are better than that utilizing only pre-trained sentence embeddings from BERT model (F1/Precision/Recall scores of 87.0/87.1/87.0 and 78.5/78.8/78.5 on PubMed 20K and NICTA-PIBOSO datasets). This indicates the role of the proposed LSTM-based network in supporting to get more comprehensive sentence representation by combining both the strength of medical domain language understanding of language model on large-scale medical corpora and task-specific features on word, character, and statistical information within specific context. 
Leveraging sentence embeddings from Sen-model, the model at the abstract level (Abs-Model) further enhance the system performance. We achieve the F1/Precision/Recall scores of 92.7/93.2/92.6 on PubMed 20K and 84.6/85.5/84.1 on NICTA-PIBOSO datasets as shown in the lower part of Table~\ref{table:res_01}. The model at the segment level (Seg-Model), when being integrated into the system, shows efficiency in considering coherent dependencies of sentences in local regions within segments and recorrect sentences at the boundary of two label classes. The best system (Combine-Model), which combines of Abs-Model and Seg-Model, achieves the best result of 92.8/93.4/92.7 on PudMed 20K and 85.3/86.5/84.5 on NICTA-PIBOSO as shown in Table~\ref{res_01_01}. This model also outperforms the baseline~\cite{bi-ann} by 1.0\%, 2.8$\%$, and 2.6$\%$ on PubMed 200K, PubMed 20K, and NICTA-PIBOSO datasets in terms of F1 scores, respectively.

Compared with the state-of-the-art systems as shown in Table~\ref{table:res_02} and Table~\ref{table:res_03}, although our best system presents fundamental network architectures at the abstract level and the segment level when leveraging the well-presented sentence embeddings at the sentence level, we achieve very competitive results (top-4 on PubMed 20K and top-3 on NICTA-PIBOSO).
This indicates that the role of the LSTM-based network (Sen-Model) at the sentence level is important to achieve comprehensive sentence representation, which can be effectively set as an initial foundation and leveraged in higher levels of segment level and abstract level to improve the model performance.
Therefore, our future work is to investigate novel methods for further improving the model performance based on the well-presented sentence representation. \vspace{-0.2cm}
\section{Conclusion}
This paper has presented a deep learning system for the Sequential Sentence Classification (SSC) task in medical scientific abstract based on the motivation of improving sentence representation. By conducting extensive experiments, we achieve the best system that outperforms the baseline by 1.0\%, 2.8\%, and 2.6\%  on the benchmark datasets of PudMed 200K RCT, PudMed 20K RCT, and NICTA-PIBOSO regarding F1 scores, respectively. The results are highly competitive to the state-of-the-art systems on these two datasets.  Particularly, our proposed LSTM-based network at the sentence level proves a vital role in generating comprehensive sentence representation, which can be served as a strong foundation for further exploring and improving the performance of the SSC task on the higher contextual levels.

\addtolength{\textheight}{-10cm}   

\section*{ACKNOWLEDGMENTS}
The work described in this paper is performed in the H2020 project STARLIGHT (“Sustainable Autonomy and Resilience for LEAs using AI against High Priority Threats”). This project has received funding from the European Union’s Horizon 2020 research and innovation program under grant agreement No 101021797.

\addtolength{\textheight}{-11cm}   

\end{document}